\documentclass[9pt,twocolumn,twoside]{optica}
\setboolean{shortarticle}{false}
\setboolean{minireview}{false}
\usepackage{url}
\usepackage{amssymb,amsfonts,amsmath}
\usepackage{times}

\title{Lensless computational imaging through deep learning}

\author[1*]{Ayan Sinha}
\author[2]{Justin Lee}
\author[1]{Shuai Li}
\author[1,3]{George Barbastathis}

\affil[1]{Department of Mechanical Engineering, Massachusetts Institute of Technology, 77 Massachusetts Avenue, Cambridge, MA 02139}
\affil[2]{Institute for Medical Engineering Science, Massachusetts Institute of Technology, 77 Massachusetts Avenue, Cambridge, MA 02139}
\affil[3]{Singapore-MIT Alliance for Research and Technology (SMART) Centre, One Create Way, Singapore 117543, Singapore}
\affil[*]{Corresponding author: sinhayan@gmail.com}

\dates{Compiled \today}

\ociscodes{(100.3190) Inverse problems; (100.4996) Pattern recognition, neural networks; (100.5070)  Phase retrieval; (110.1758)  Computational imaging.}

\doi{\url{http://dx.doi.org/10.1364/optica.XX.XXXXXX}}

\begin{abstract}
Deep learning has been proven to yield reliably generalizable answers to numerous classification and decision tasks. Here, we demonstrate for the first time, to our knowledge, that deep neural networks (DNNs) can be trained to solve inverse problems in computational imaging. We experimentally demonstrate a lens-less imaging system where a DNN was trained to recover a phase object given a  raw intensity image recorded some distance away.
\end{abstract}

\setboolean{displaycopyright}{true}

\begin{document}

\maketitle
\thispagestyle{fancy}
\ifthenelse{\boolean{shortarticle}}{\abscontent}{}

\section{Introduction}

Neural network training can be thought of as generic function approximation: given a training set ({\it i.e.}, examples of matched input and output data obtained from a hitherto-unknown model), a neural network attempts to generate a computational architecture that accurately maps all inputs in a test set (distinct from the training set) to their corresponding outputs. In this paper, we propose that deep neural networks may ``learn'' to approximate solutions to inverse problems in computational imaging.

A general computational imaging system consists of a physical part and computational part. In the physical part, light propagates through one or more objects of interest as well as optical elements such as lenses, prisms, etc. finally producing a raw intensity image on a digital camera. The raw intensity image is then computationally processed to yield object attributes, e.g. a spatial map of light attenuation and/or phase delay through the object---what we traditionally call an ``intensity image'' or ``quantitative phase image,'' respectively. The computational part of the system is then said to have produced a solution to the inverse problem.

The study of inverse problems is traced back at least a century to Tikhonov \cite{tikhonov} and Wiener \cite{wiener}. A good introductory book with rigorous but not overwhelming discussion of the underlying mathematical concepts, especially regularization, is \cite{bertero}. During the past decade, the field experienced a renaissance due to the almost simultaneous maturation of two related mathematics disciplines: convex optimization and harmonic analysis, especially sparse representations. A light technical introduction to these fascinating developments can be found in \cite{candes2008} and a more detailed exposition in \cite{bradybook}.

Neural networks have their own history of legendary ups-and-downs \cite{minsky} culminating with an even more recent renaissance. This was driven by empirical findings that deep multi-layer architectures, dubbed as ``deep neural networks'' (DNNs), could generalize better than had been previously thought possible. Vast improvements in the available computational power were certainly helpful; most effective, however, were revivals of older concepts combined with new insights on these concepts' function and realization. These have included: architectures, such as convolutional connectivity \cite{lecun1989,hinton2012,Szegedy2015,He2016} for regularization and pruning; nonlinearities, such the now widespread use of non-differentiable piecewise linear units \cite{fukushima75} as opposed to the older sigmoidal functions that were differentiable but also prone to stagnation \cite{goodfellow16book}; and algorithms, such as more efficient backprop \cite{rumel89,lecun10}. Within the last four-five years, neural networks have exhibited spectacular success at solving ``hard'' computational problems: playing complex games like Atari \cite{Mnih2015} and Go \cite{Silver2016}; object generation \cite{dosovitskiy2014}; object detection \cite{lecun2015}; and image restoration: colorization \cite{cheng2015}, deblurring \cite{dong2014,xu2014,sun2015}, and in-painting \cite{xie2012}.

\begin{figure}[h]
\centering\includegraphics[width=0.95\linewidth]{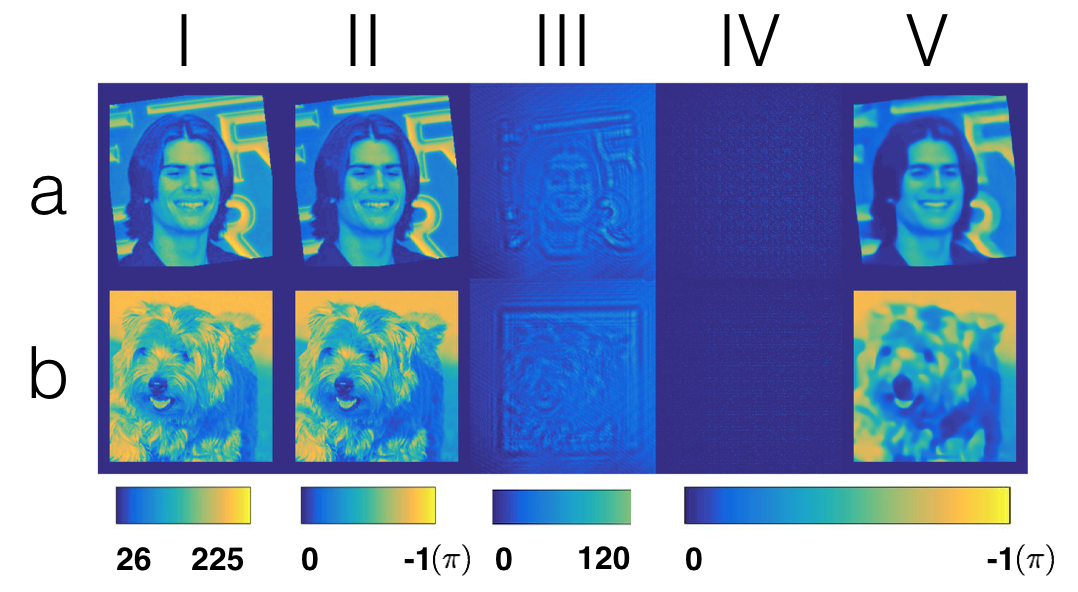}
\caption{DNN training. Rows (a) and (b) denote the networks trained on Faces-LFW  and ImageNet dataset, respectively. (i) randomly selected example drawn from the database; (ii) calibrated phase image of the drawn sample; (iii) diffraction pattern generated on the CMOS by the same sample; (iv) DNN output before training ({\it i.e.} with randomly initialized weights); (v) DNN output after training.}
\label{fig:FEXPTRAIN}
\end{figure}

The hypothesis that we set out to test in this paper is whether a neural network can be trained to recover object estimates from raw intensity images (i.e. solve the inverse problem). This is a rather general question and may take several flavors, depending on the nature of the object, the physical design of the imaging system, etc. We chose to test our hypothesis in a very specific ``heavy'' computational imaging scenario: a lensless optical setup where diffraction patterns of pure phase objects under coherent illumination were captured as ``raw images''.

Our experimental arrangement, described in more detail in Section~\ref{sec:experiment}, falls in-between two categories of imaging systems that could be traditionally called "digital holographic imaging,'' \cite{kim2010} and ``transport-of-intensity imaging'' \cite{Teague1983, Streibl1984}. It is neither, because it violates the necessary assumptions of sparse objects leaving most of the incoming light unscattered to serve as reference beam for the digital hologram; and of sparse object gradients that avoid singularities in the transport-of-intensity equation. Hence, either technique would be expected to require significant fine-tuning of regularization parameters to yield satisfactory results.

The idea of using neural networks to clean up images isn't exactly new. For example, Hopfield's associative memory network \cite{hopfield} was capable of retrieving entire faces from partially obscured inputs, and was implemented in an all-optical architecture \cite{hopfieldoptical} when computers weren't nearly as powerful as they are now. Recently, Horisaki et al. \cite{Horisaki2016} used support-vector machines, a form of bi-layer neural network with nonlinear discriminant functions, also to recover face images when the obscuration is caused by scattering media. Here, we extend upon the aforementioned efforts and train a deep neural network to recover images of objects given ``raw image'' measurements of the modulus of their diffraction patterns.

Our results demonstrate that DNNs are capable of ``learning'' the inverse mapping between raw intensity image and object {\em directly} from experimental data. Our results also suggest that the neural network ``learns'' the underlying governing equations of the system, including its forward operator and possible deviations from underlying idealizations and assumptions. This lack of need for a prior model is notable because it removes the difficulty of correctly specifying the forward operator; many optimization approaches are sensitive to errors due to inaccurate or incomplete forward models.

\begin{figure}[t]
\centering\includegraphics[width=0.8\linewidth]{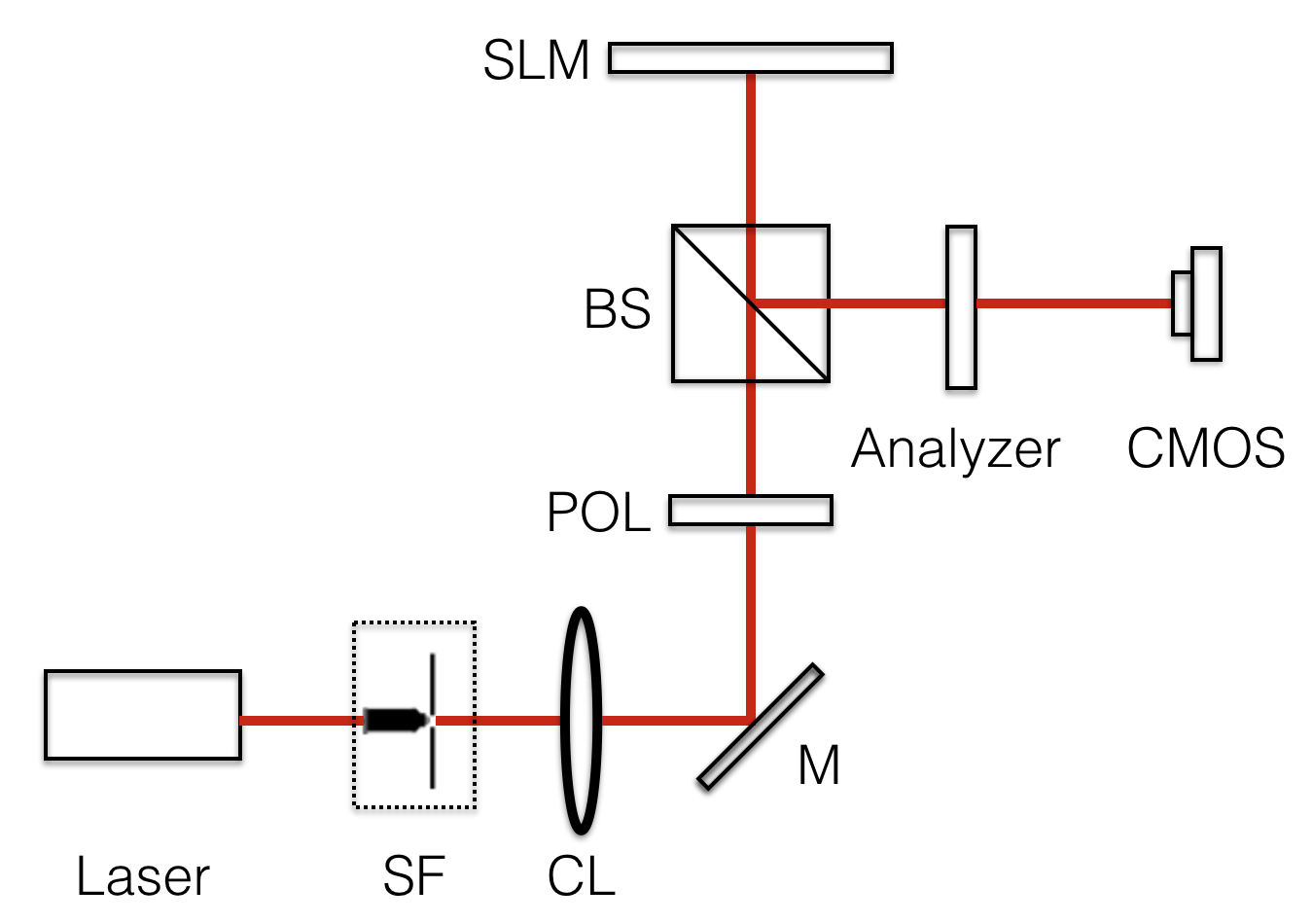}
\caption{Experimental arrangement. SF: spatial filter; CL: collimating lens; M: mirror; POL: linear polarizer; BS: beam splitter; SLM: spatial light modulator.}
\label{fig:FEXPSET}
\end{figure}

Neural network approaches often come under criticism because the quality of training depends on the quality of the examples given to the network during the training phase. For instance, if the inputs used to train a network are not diverse enough, then the DNN will learn priors of the input images instead of generalized rules for ``cleaning up images.'' This was the case in \cite{Horisaki2016}, where an SVM trained using images of faces could adequately reconstruct faces, but when given the task of reconstructing images of natural objects such as a pair of scissors, the trained SVM still returned an output that resembled a human face.

For our specific problem, an ideal training set would encompass all possible ``phase objects.'' Unfortunately, ``phase objects,'' generally speaking, constitute a rather large class of objects and it would be unrealistic to attempt to train a network sampling from across all possible objects from this large class. Instead, we synthesize phase objects in the form of natural images derived from the ImageNet \cite{imagenet} database because it is readily available and widely used in the study of various machine learning problems. For comparison, we also trained a separate network using a narrower class of (facial) images from the Faces-LFW \cite{faceslfw} database.

As expected, our network did well when presented with unknown phase objects in the form of faces or natural images that it had been trained to. Notably, the network also performed well when presented with objects outside of its ``training class'' -- the DNN trained using images of faces was able to reconstruct images of natural objects, and the DNN trained using images of natural objects was able to reconstruct images of faces. Additionally, both DNNs were able to reconstruct completely distinct images including: handwritten digits, characters from different languages (Arabic, Mandarin, English), and images from a disjoint natural image dataset.

Both trained networks yielded accurate results even when the object-to-sensor distance(s) in the training set slightly differed from that of the testing set, suggesting that the network is not merely pattern-matching but instead has actually ``learned'' a generalizable model approximating the underlying system.

The details of our experiment, including the physical system and the computational training and testing results, are described in Section~\ref{sec:experiment}. The neural network itself is analyzed in Section~\ref{sec:analysis}, and concluding thoughts are in Section~\ref{sec:conclusions}.

\begin{figure}[h]
\centering\includegraphics[width=0.99\linewidth]{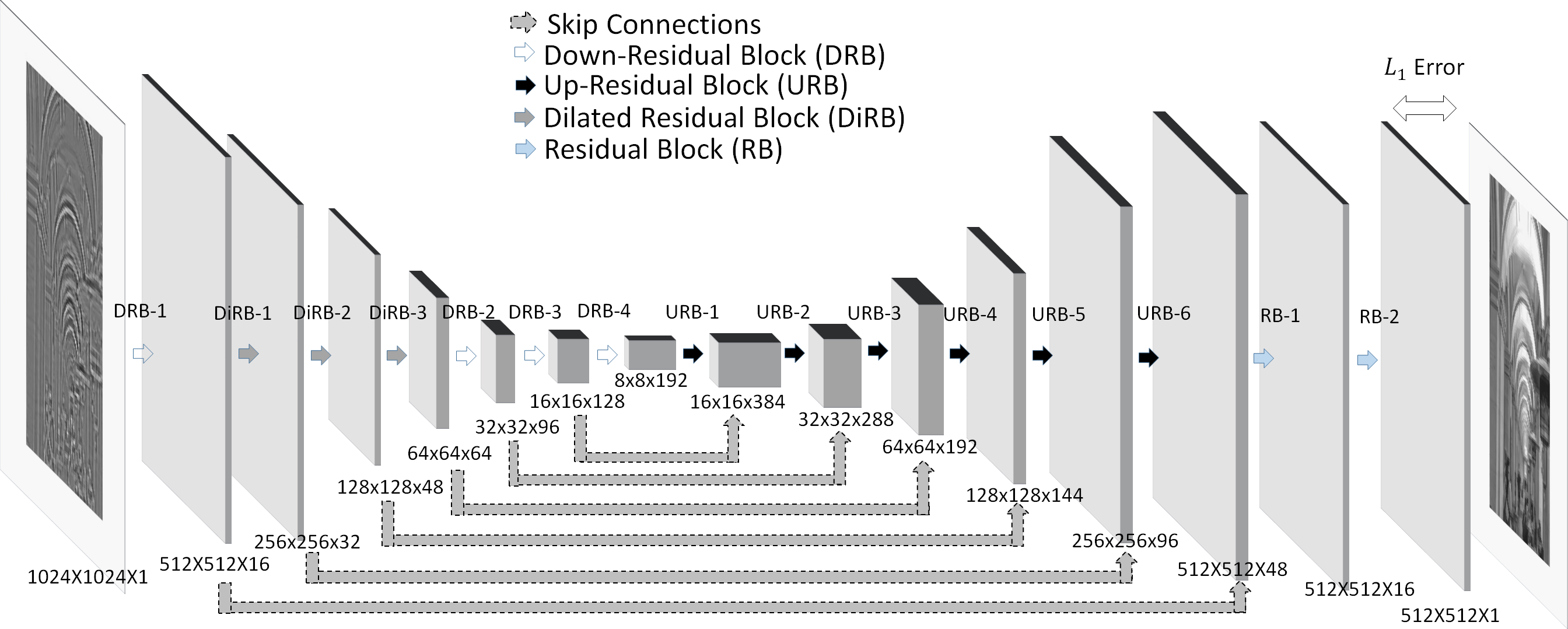}
\caption{Detailed schematic of our DNN architecture, indicating the number of layers, nodes in each layer, etc. }
\label{fig:FNN}
\end{figure}

\section{Experiment } \label{sec:experiment}

Our experimental arrangement is as shown in Figure \ref{fig:FEXPSET}. Light from a He-Ne laser source (Thorlabs, HNL210L, 632.8nm) first transmits through a spatial filter, which consists of a microscope objective (Newport, M-60X, 0.85NA) and a pinhole aperture ($D=5\mu \text{m}$), to remove spatial noise. After being collimated by the lens ($f=150$mm), the light is reflected by a mirror and then passes through a linear polarizer to set the appropriate polarization. After that, the light is split by a beam splitter. A spatial light modulator (Holoeye, LC-R 720, reflective) is placed normally incident to the transmitted light and acts as a pixel-wise phase object. The SLM-modulated light is then reflected by the beam splitter and passes through a linear polarization analyzer, before being collected by a CMOS camera (Basler, A504k). Images recorded are then processed on an Intel i7 CPU, with neural network computations performed on a GTX1080 graphics card (NVIDIA).

\begin{figure*}[h]
\centering\includegraphics[width=0.86\linewidth]{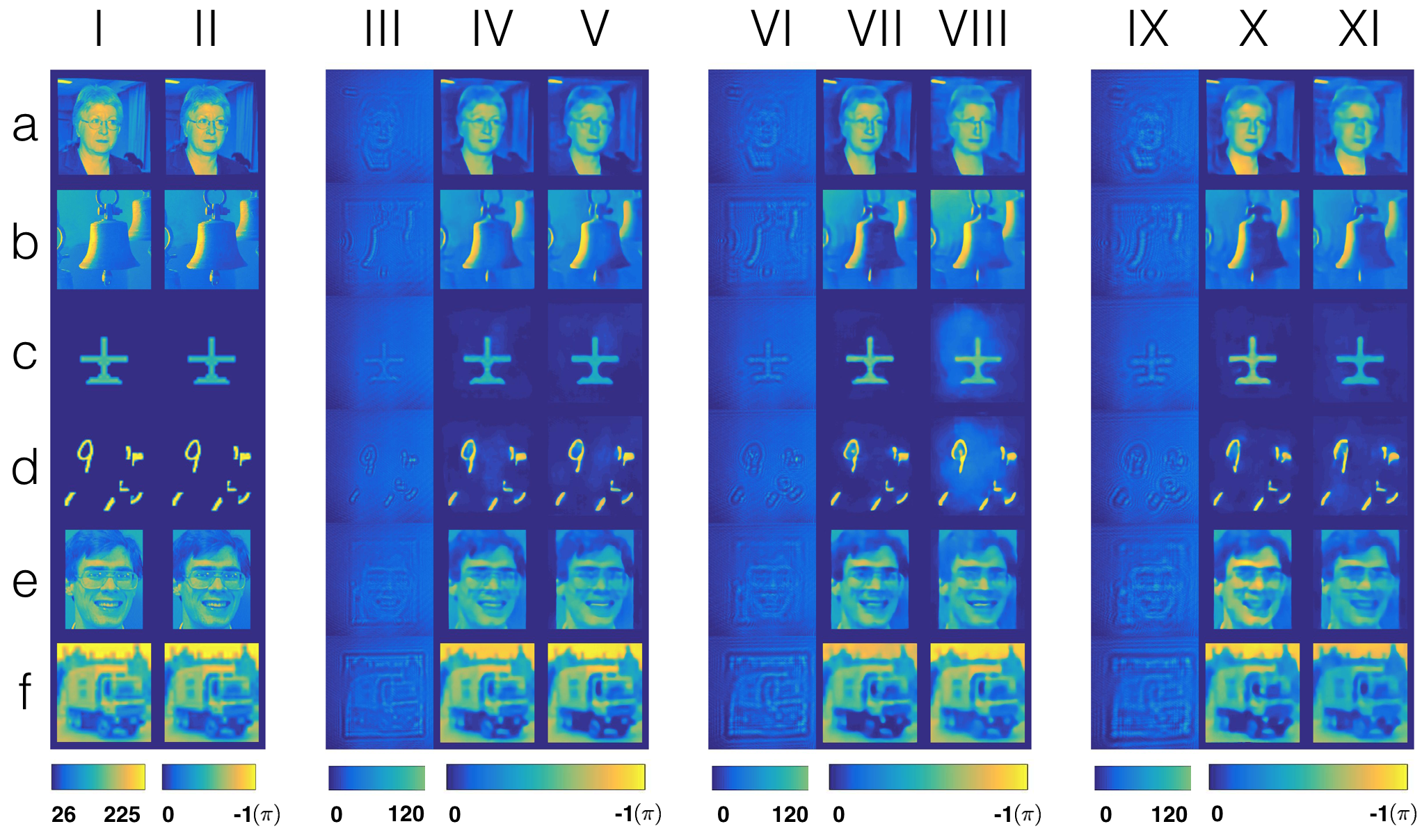}
\caption{Qualitative analysis of our trained deep neural networks for three object-to-sensor distances (37.5 cm, 67.5 cm and 97.5 cm) on different datasets.
(i) Ground truth pixel value inputs to the SLM. (ii) Corresponding phase imaged calibrated by SLM response curve. (iii) Raw intensity images captured by CMOS detector at distance $d \sim37.5 cm$. (iv) DNN reconstruction from raw images when trained using Faces-LFW dataset. (v) DNN reconstruction when trained used ImageNet dataset. Columns (vi-viii) and (ix-xi) follow the same sequence as (iii-v) but in these sets the CMOS is placed at a distance of $\sim67.5 cm$ and $\sim97.5 cm$, respectively.  Rows (a-f) correspond to the dataset from which the test image is drawn: (a) Faces-LFW, (b) ImageNet, (c) Characters, (d) MNIST Digits, (e) Faces-ATT, or (f) CIFAR. }
\label{fig:FEXPTEST}
\end{figure*}

According to its user manual, the LC-R 720 SLM can realize (approximate) pure-phase modulation if we modulate the light polarization properly. Specifically, for He-Ne laser light, if we set the polarization of the incident beam at $45^{\circ}$ linearly polarized with respect to the vertical direction and also set the linear polarization analyzer to be oriented at $340^{\circ}$ with respect to the vertical direction, then the amplitude modulation of the SLM will become almost independent of the assigned (8-bit gray-level) input. In this arrangement, the phase modulation of the SLM follows a monotonic, almost-linear relationship with the assigned pixel value (with maximum phase depth: $\sim1\pi$). We experimentally evaluated the correspondence between 8-bit grayscale input images projected onto the SLM and phase values in the range $[0,-\pi]$ (see supplement). In this paper, we approximate our SLM as a pure-phase object and computationally recover the phase using a neural network.



The CMOS detector was placed after a free-space propagation distance $d$, which ranged from $\sim37.5-97.5 cm$ to record diffraction patterns. Our experiment consists of two phases: training and testing. During the training phase, we modulate the phase SLM according to samples randomly selected from the Faces-LFW or ImageNet database. We resize, and pad selected images before displaying them on our SLM.  Two examples of inputs, as they are sent to the SLM, and their corresponding raw intensity images (diffraction patterns) as captured on the CMOS are shown in Figure \ref{fig:FEXPTRAIN}. Our training set consisted of 10,000 such faces/images - diffraction pattern pairs. The raw intensity images from all these training examples were used to train the weights in our DNN. We used a Zaber A-LST1000D stage with repeatability 2.5$\mu m$ to translate the camera in order to analyze the robustness of the learnt network to perturbations (See: Network Analysis).

\begin{figure}[th]
\centering\includegraphics[width=\linewidth]{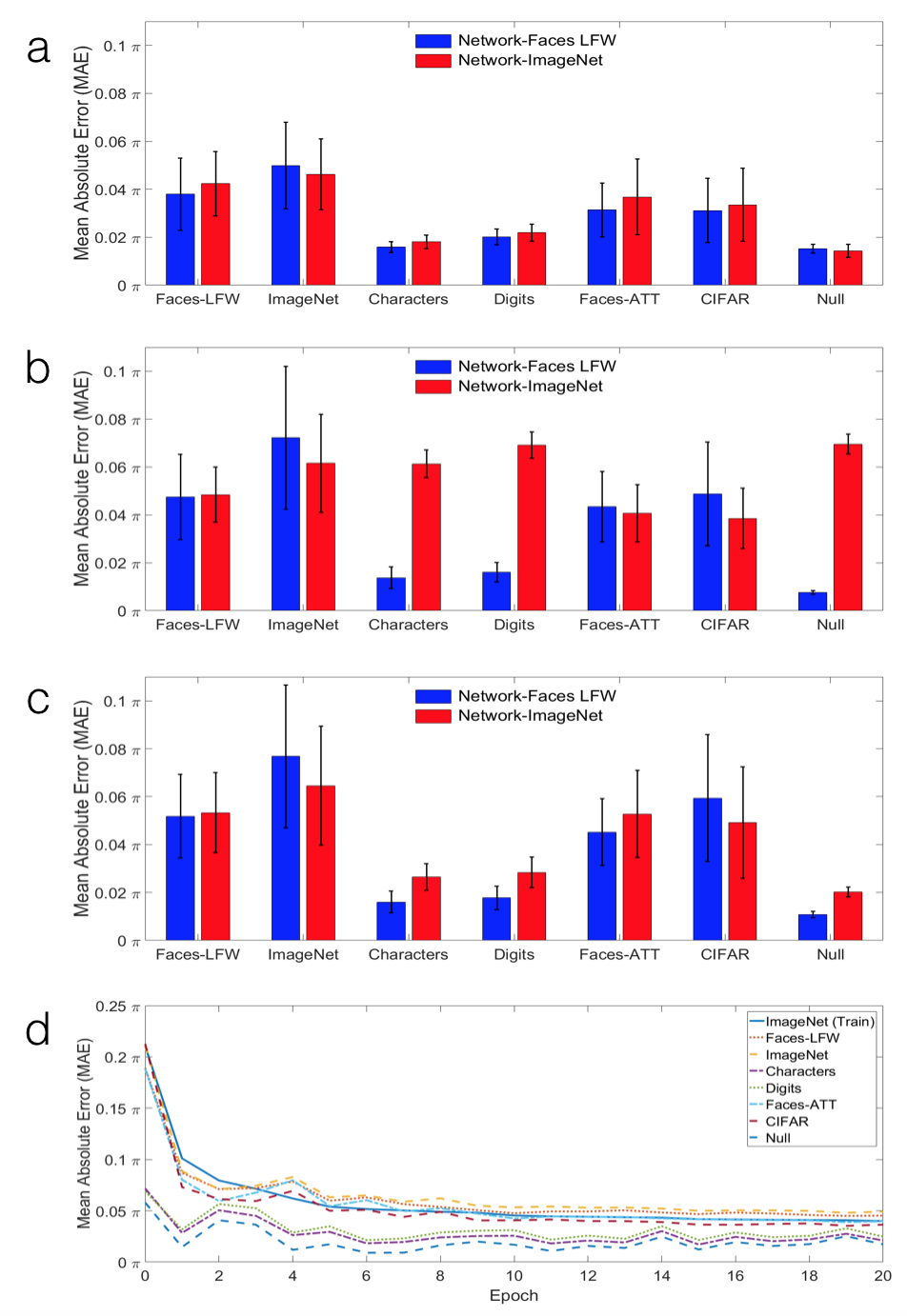}
\caption{Quantitative analysis of  our trained deep neural networks for 3 object-to-sensor distances of (a) 37.5 cm, (b) 67.5 cm, and (c) 97.5 cm for the DNNs trained on Faces-LFW (blue) and ImageNet (red) on 7 datasets. (d) The training and testing error curves for network trained on ImageNet at distance 37.5 cm over 20 epochs.}
\label{fig:FERRORPLOTS}
\end{figure}

Our DNN uses a convolutional residual neural network (ResNet) architecture. In a convolutional neural network (CNN), inputs are passed from nodes of each layer to the next, with adjacent layers connected by convolution. Convolutional ResNets extend CNNs by adding short term memory to each layer of the network. The intuition behind ResNets is that one should only increase the depth of the neural network if he/she stands to gain something by adding that extra layer. The intuition behind ResNets is that one should only adds a new layer if you can get something extra out of adding that layer. ResNets ensure that the $N+1$th layer learns something new about the network by also providing the original input to the output of the $(N+1)$th layer and performing calculations on the residual of the two. This forces the new layer to learn something different from what the input has already encoded/learned \cite{He2016}.

A diagram of our specific DNN architecture is shown in Fig. \ref{fig:FNN}. The input layer is the image captured by the CMOS camera. It is then successively decimated by 7 residual blocks of convolution + downsampling followed by 6 residual blocks of deconvolution + upsampling, and finally 2 standard residual blocks. Some of the residual blocks are comprised of dilated convolutions so as to increase the receptive field of the convolution filters, and hence, aggregate diffraction effects over multiple scales \cite{Koltun}. We use skip connections to pass high frequency information learnt in the initial layers down the network towards the output reconstruction, and  have two standard residual blocks at the end of the network to finetune the reconstruction. At the very last layer of our CNN, the values represent an estimate of our input signal. The connection weights are trained using backpropagation (not to be confused with optical backpropagation) on the $L_1$ error between the network output and the nominal appearance of the training samples represented as:
\begin{equation} \label{error}
\vspace{-2mm}
min \frac{1}{wh}\sum_{(m,n)} \parallel (Y(m,n)-G(m,n)) \parallel_1
\end{equation}
Here, $w,h$ are the width and the height of the output, $Y$ is the output of the last layer, and $G$ is the ground truth phase value. $G(m,n)$ lies in the range $[0,-\pi]$.

We collected data from six separate experiment runs using training inputs from Faces-LFW or ImageNet and object-to-sensor distances of $37.5 cm$, $67.5 cm $ , or $97.5 cm$. These data were used to train six separate DNNs for evaluation.

Fig. \ref{fig:FEXPTRAIN}(iv) shows a sample DNN's output at the beginning of its training phase ({\it i.e.} with randomly initialized weights), and Fig. \ref{fig:FEXPTRAIN}(v) shows the network output after training, for the same example object-raw image pairs. Training each network took $\approx 20$ hours using Tensorflow and a Nvidia GTX 1080 GPU. We provide analysis of the trained DNN in Section~\ref{sec:analysis}.

Our testing phase consisted of: (1) sampling disjoint examples from the same database (either Faces-LFW or ImageNet) and other databases such as MNIST, CIFAR, Faces-ATT etc., (2) using these test examples to modulate the SLM and produce raw intensity images on the camera, (3) passing these intensity images as inputs to out trained DNN, and (4) comparing the output to ground truth.

\begin{figure}[h!]
\centering\includegraphics[width=0.9\linewidth]{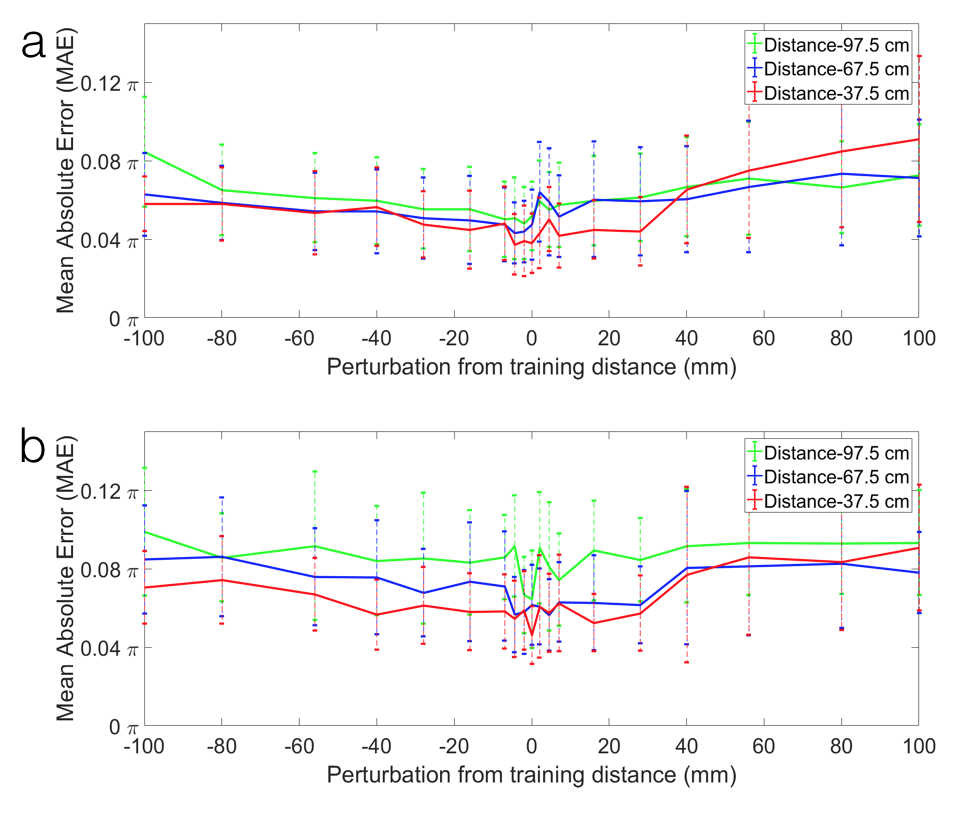}
\caption{Quantitative analysis of the sensitivity of the trained deep convolutional neural network to the object-to-sensor distance. The network was trained on (a) Faces-LFW database and (b) ImageNet and tested on disjoint Faces-LFW and ImageNet sets, respectively.}
\label{fig:FINSIDE2}
\end{figure}


\begin{figure}[h!]
\centering\includegraphics[width=0.9\linewidth]{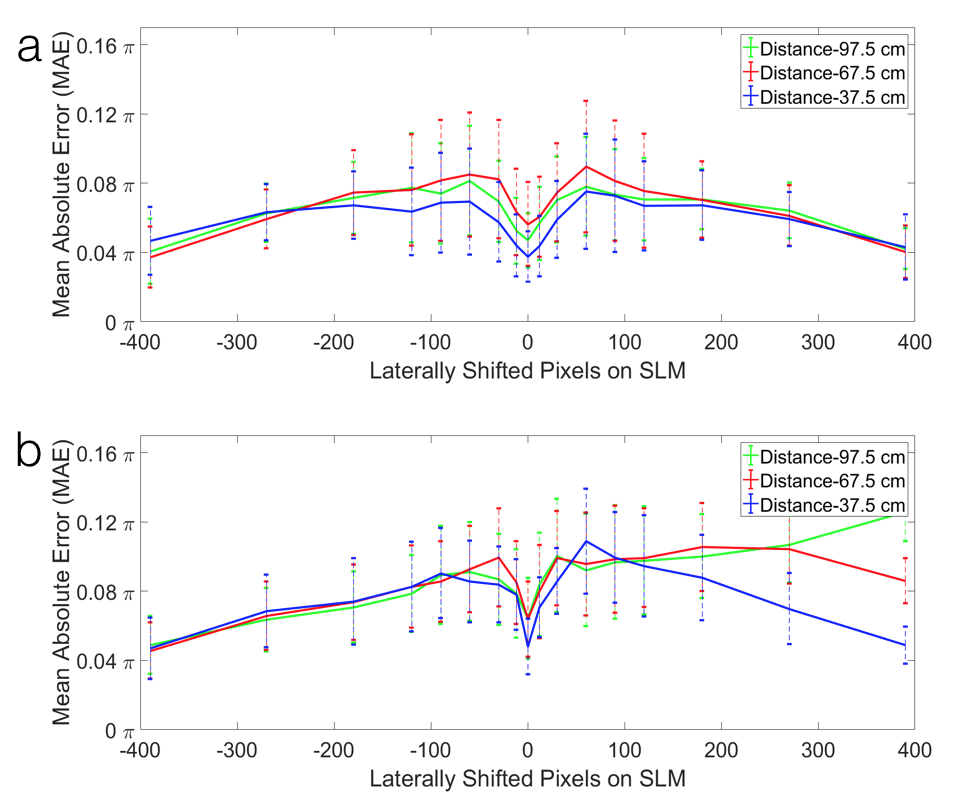}
\caption{Quantitative analysis of the sensitivity of the trained deep convolutional neural network to laterally shifted images on the SLM. The network was trained on (a) Faces-LFW database, (b) ImageNet and tested on disjoint Faces-LFW and ImageNet sets, respectively.}
\label{fig:FINSIDE4}
\end{figure}

\begin{figure}[h!]
\centering\includegraphics[width=0.8\linewidth]{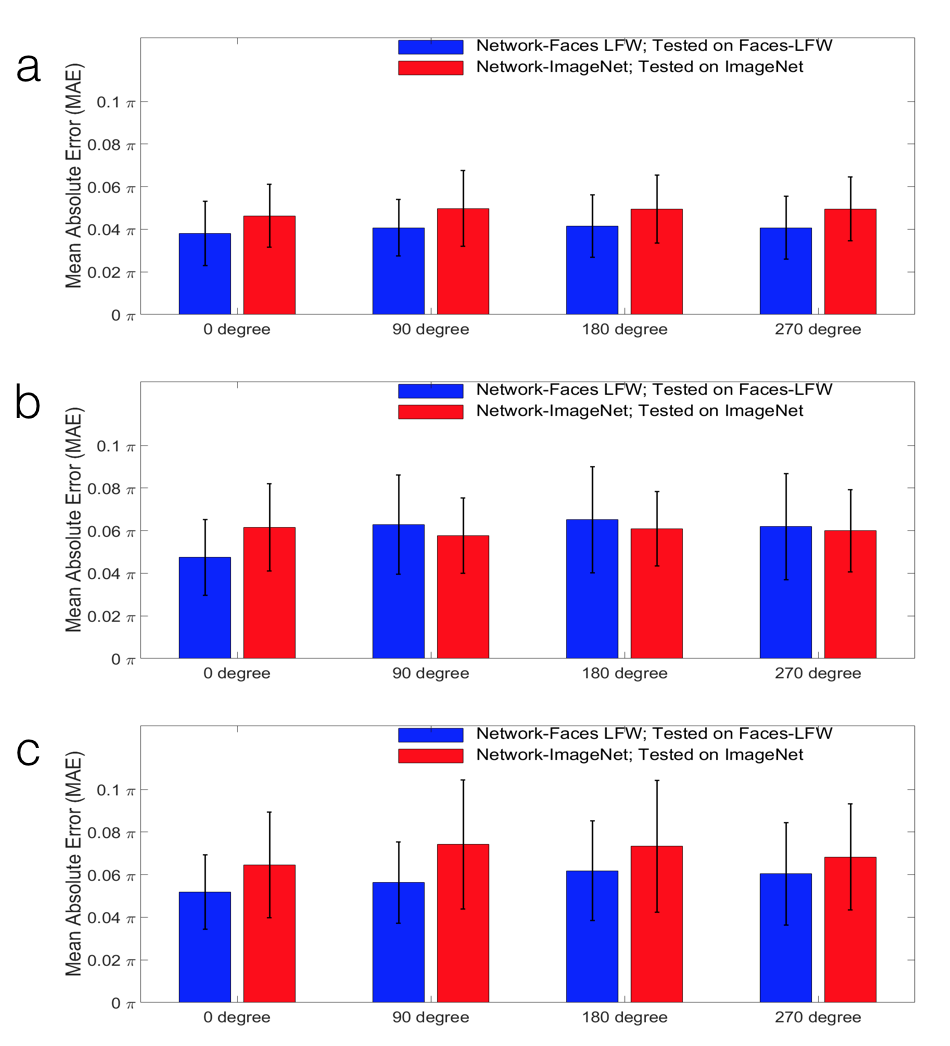}
\caption{Quantitative analysis of the sensitivity of the trained deep convolutional neural network to rotation of images on the SLM.  The baseline distance on which the network was trained is (a) 37.5 cm, (b) 67.5 cm and (c) 97.5 cm, respectively. }
\label{fig:FINSIDE5}
\end{figure}

\begin{figure}[h]
\centering\includegraphics[width=\linewidth]{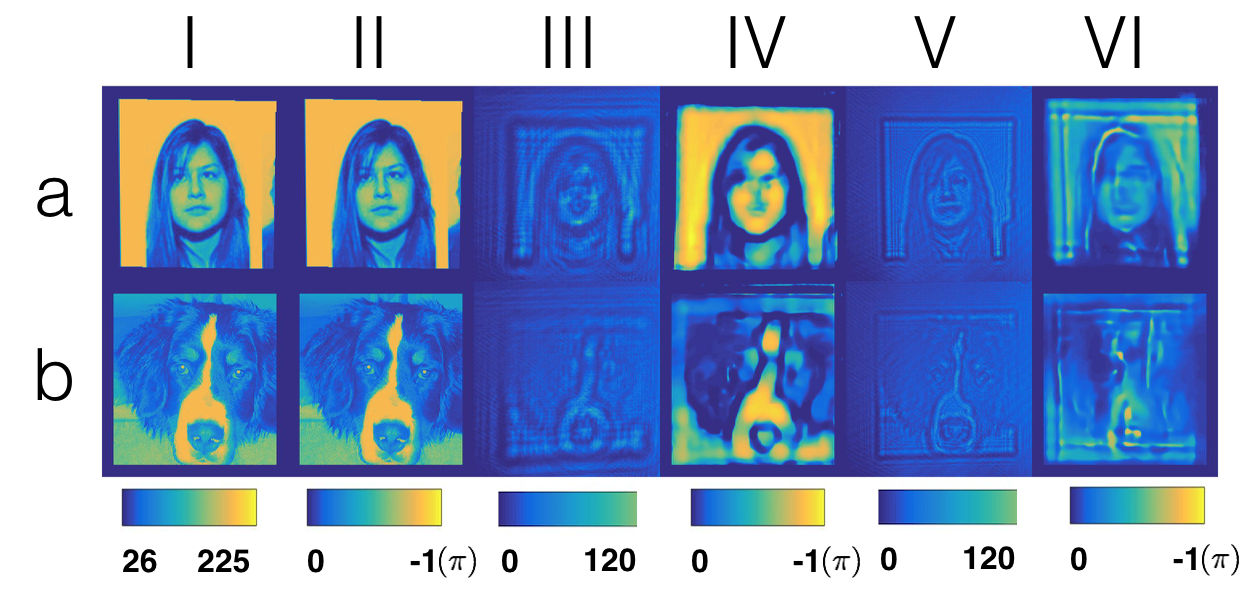}
\caption{Failure cases on networks trained on Faces-LFW (row $a$) and ImageNet (row $b$) datasets. (i) Ground truth input, (ii) calibrated phase input to SLM, (iii) raw image on camera (iv) reconstruction by DNN trained on images at distance 37.5 cm between SLM and camera and tested on images at distance 107.5 cm, (v) raw image on camera and (vi) reconstruction by network trained on images at distance 97.5 cm between SLM and camera and tested on images at distance 27.5 cm.}
\label{fig:Fperturb5}
\end{figure}

\begin{figure}[h!]
\centering\includegraphics[width=0.99\linewidth]{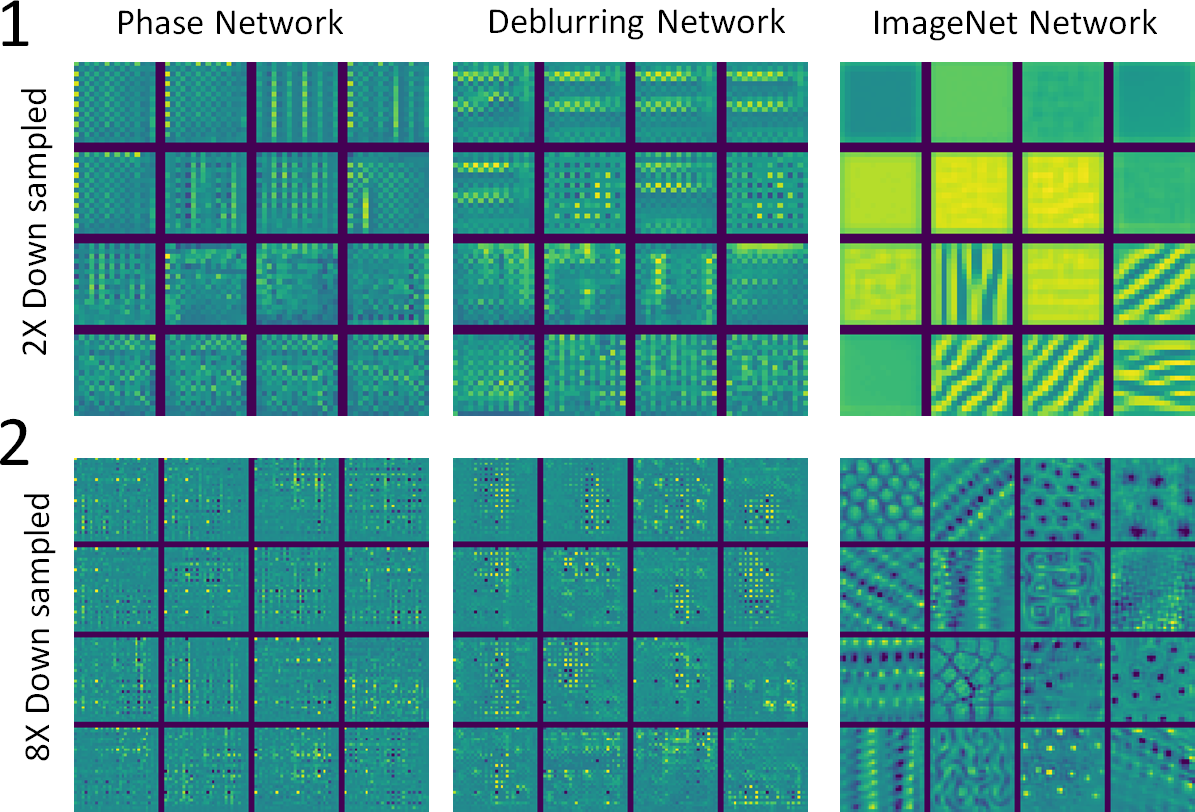}
\caption{(1) $16 \times 16$ inputs that maximally activate the last set of 16 convolutional filters in layer 1 of our phase retrieval network trained on ImageNet at distance of 37.5 cm, a deblurring network, and an ImageNet classification network. The image is downsampled by a factor of 2 in this layer. (2) $32 \times 32$ inputs that maximally activate the last set of 16 randomly chosen convolutional filters in layer 3 of: our network, a deblurring network, and ImageNet classification network. The raw image is downsampled by a factor of 8 in this layer.}
\label{fig:FINSIDE}
\end{figure}

\begin{figure*}[h]
\centering\includegraphics[width=0.7\linewidth]{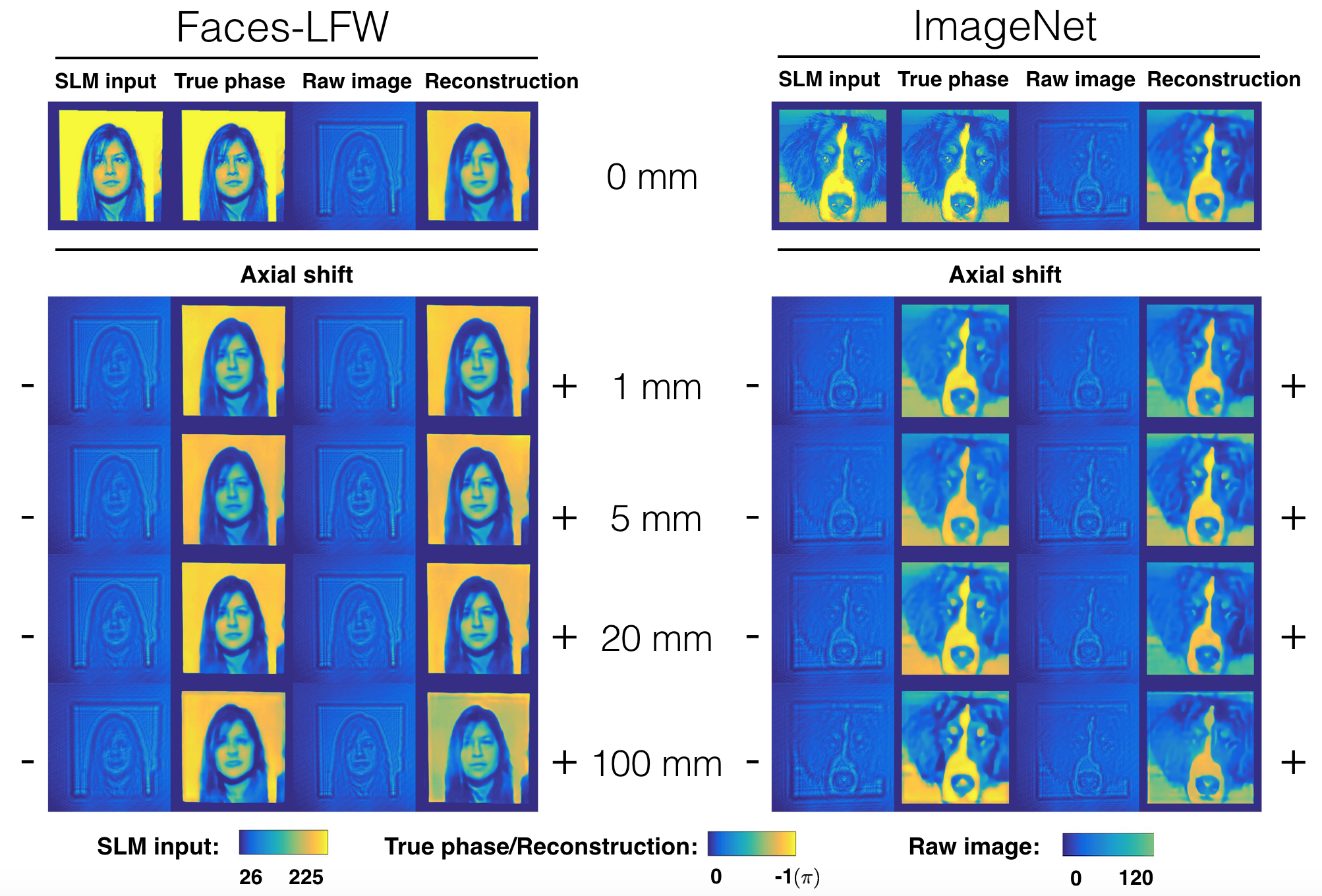}
\caption{Qualitative analysis of the sensitivity of the trained deep convolutional neural network to the object-to-sensor distance. The baseline distance on which the network was trained is 37.5 cm. }
\label{fig:Fperturb1}
\end{figure*}


\begin{figure*}[h]
\centering\includegraphics[width=0.7\linewidth]{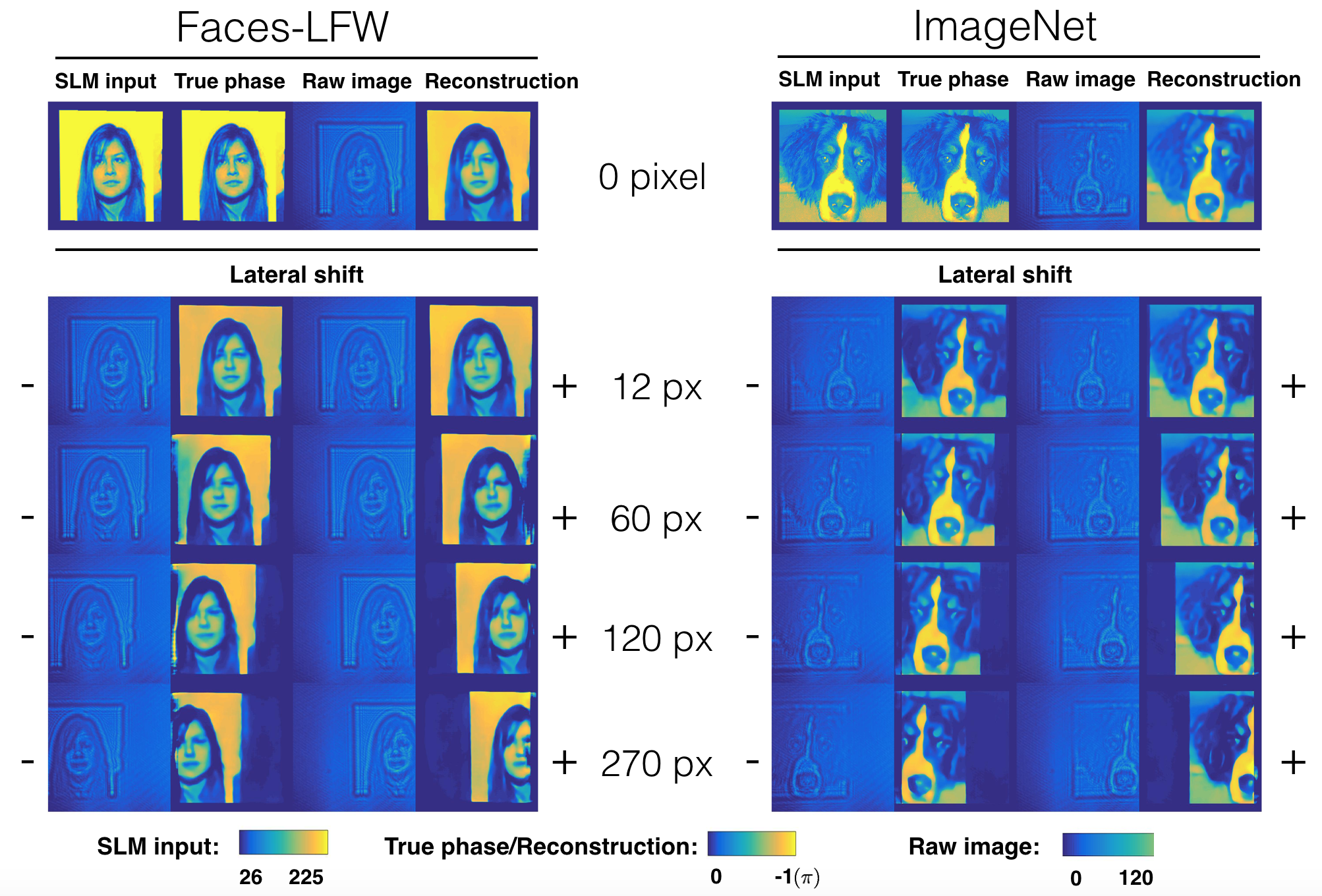}
\caption{Qualitative analysis of the sensitivity of the trained deep convolutional neural network to lateral shifts of images on the SLM. The baseline distance on which the network was trained is 37.5 cm. }
\label{fig:Fperturb3}
\end{figure*}

\begin{figure*}[h]
\centering\includegraphics[width=0.7\linewidth]{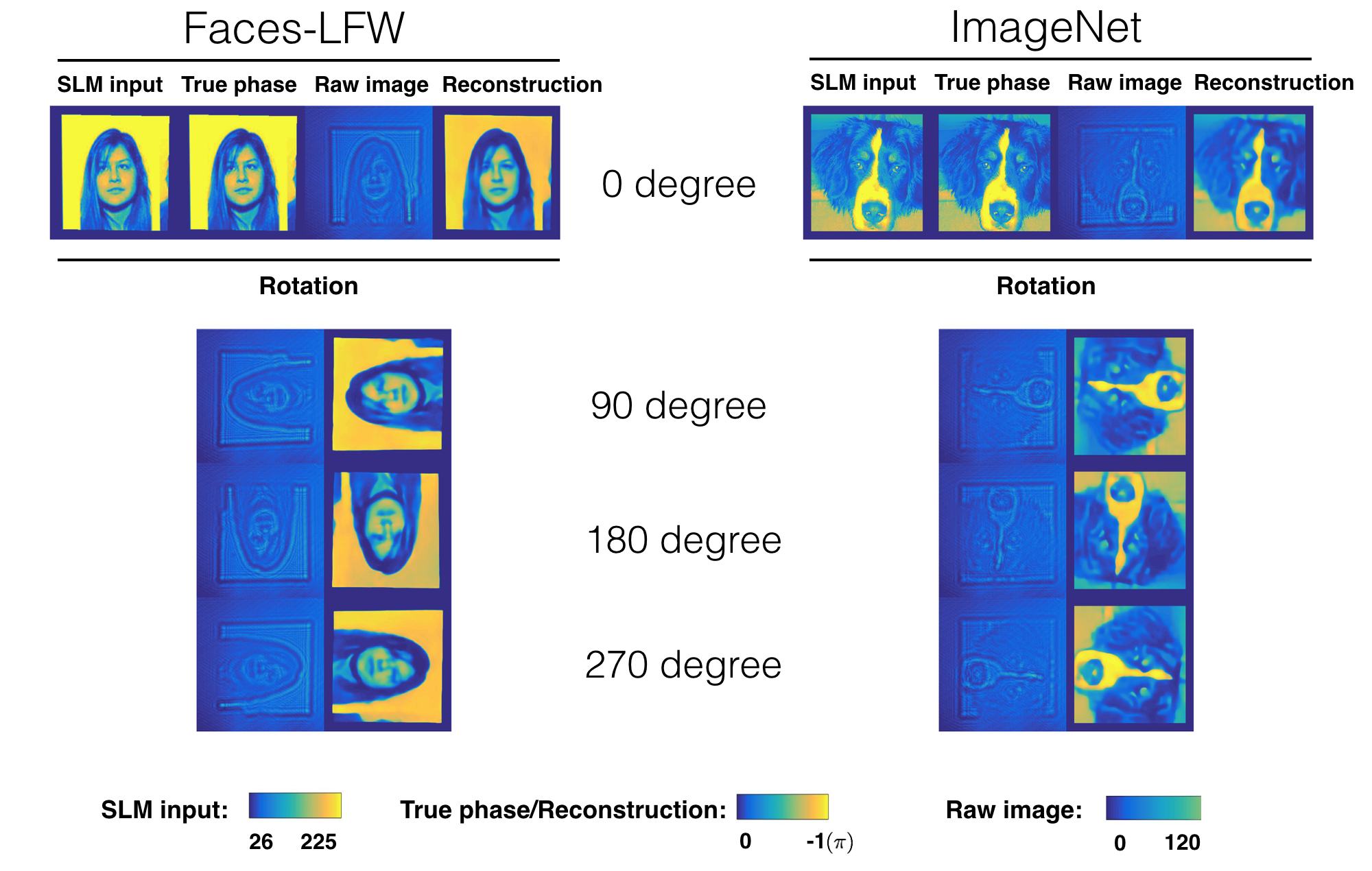}
\caption{Qualitative analysis of the sensitivity of the trained deep convolutional neural network to rotation of images in steps of 90. The baseline distance on which the network was trained is 37.5 cm. }
\label{fig:Fperturb4}
\end{figure*}

\section{Results and Network analysis} \label{sec:analysis}

The standard method of characterizing neural network training is by plotting the progression of training and test error across training epochs (iterations in the backpropagation algorithm over all examples). These curves are shown in Figure \ref{fig:FERRORPLOTS} for our network trained using the ImageNet database and tested using images from: (a) Faces-LFW (b) a disjoint ImageNet set, (c) images from an English/Chinese/Arabic characters database, (d) the MNIST handwritten digit database, (e) Faces-ATT,  (f) CIFAR, (g) a constant-value "Null" image. Our ImageNet learning curves in Figure \ref{fig:FERRORPLOTS}d show convergence to low value after $\sim$10 epochs, indicating that our network has not overfit to our training dataset. We plot bar graphs for the mean absolute error (MAE) over test examples in the 7 different datasets for each of the 3 object-to-sensor distances in Figure \ref{fig:FERRORPLOTS}. Lower MAE was reported for test images with large patches of constant value (characters, digits, Null) as their sparse diffraction patterns were easier for our DNN to invert. Notably, both our bar graphs and learning curves show low test error for the non-trained images, suggesting that our network generalizes well across different domains.


This is an important point and worth emphasizing: despite the fact that our network was trained exclusively on images from the ImageNet database -- i.e., images of planes, trains, cars, frogs, artichokes, etc., it is still able to accurately reconstruct images of a completely different class (e.g., faces, handwritten digits, and characters from different languages). This strongly suggests that our network has learned a model of the underlying physics of the imaging system or at the very least a generalizable mapping of low-level textures between our output diffraction patterns and input images.

A more pronounced qualitative example demonstrating this is shown in the columns (iv) (vii) and (x) of Figure \ref{fig:FEXPTEST}. Here, we trained our network using images exclusively from the Faces-ATT database. Despite this limited training set, the learned network was able to accurately reconstruct images from the ImageNet, handwritten digits, and characters datasets. This is in contrast to results shown in \cite{Horisaki2016}, where an SVM trained on images of faces was able to accurately reconstruct images of faces but not other classes of objects.

How robust is our network to sensor displacement? Is it shift and rotation invariant? To answer these questions, we fed our trained network raw intensity images at different lateral and axial positions, relative to that of the training set images. Quantitative results of these perturbations are shown in Figures \ref{fig:FINSIDE2},  \ref{fig:FINSIDE4}, \ref{fig:FINSIDE5}, and qualitative results for the networks trained at distance 37.5 cm are shown in Figures \ref{fig:Fperturb1}, \ref{fig:Fperturb3} and \ref{fig:Fperturb4}. Qualitative results for the other 2 distances are in the supplement. The results show that our trained network is robust to moderate perturbations in sensor displacement and is somewhat shift and rotation invariant. As expected, the system fails when the displacement is significantly greater (Figure \ref{fig:Fperturb5}).

%

What exactly is our network learning? To get a sense of what the network has learned, we examined its maximally-activated patterns (MAPs), i.e., what types of inputs would maximize network filter response (gradient descent on the input with average filter response as loss function \cite{zeiler2015}). Our results are shown in Figure \ref{fig:FINSIDE} and compared with the results of analogous analysis of a de-blurring network of similar architecture as well as an ImageNet classification DNN. Compared with MAPs of ImageNet and a Deblurring network, the MAPs of our phase-retrieval network show much finer/low-level textures at deep layers in the network. This suggests that the network is utilizing low-level textures (representative of a wide variety of localized diffraction patterns) when learning how to invert our inverse problem.

\section{Conclusions and discussion} \label{sec:conclusions}

The architecture presented here was deliberately well controlled, with an SLM creating the phase object inputs to the neural network for both training and testing. This allowed us to quantitatively and precisely analyze the behavior of the learning process. Application-specific training, e.g. replacing the SLM with physical phase objects for more practical applications, we judged beyond the scope of the present work. Other obvious and useful extensions would be to include optics, e.g. a microscope objective for microscopic imaging in the same mode; and to attempt to reconstruct complex objects, {\it i.e.} imparting both attenuation and phase delay to the incident light. The significant anticipated benefit in the latter case is that it would be unnecessary to characterize the optics for the formulation of the forward operator---the neural network should ``learn'' this automatically as well.  We intend to undertake such studies in future work.

\section*{Funding Information}
 This research was funded by the Singapore National Research Foundation through the SMART program (Singapore-MIT Alliance for Research and Technology) and by the Information Advanced Research Projects Agency (iARPA) through the RAVEN Program. Justin Lee acknowledges funding from the U.S. Department of Energy Computational Science Graduate Fellowship (CSGF) (DE-FG02-97ER25308).
\section*{Acknowledgments}

We gratefully acknowledge Ons M’Saad for help with the experiments, and Petros Koumoutsakos and Zhengyun Zhang for useful discussions and suggestions.

\noindent
See Supplement for supporting content.

%
%
%
%
%
%
%


\ifthenelse{\boolean{shortarticle}}{%
\clearpage
\bibliographyfullrefs{sample}
}{}


\end{document}